\documentclass[10pt,journal,compsoc]{IEEEtran}
\ifCLASSOPTIONcompsoc
  \usepackage[nocompress]{cite}
\else
  \usepackage{cite}
\fi
\ifCLASSINFOpdf
\else
\fi

\usepackage{times}
\usepackage{color}
\usepackage{url}
\usepackage{graphicx}
\usepackage{array}
\usepackage{colortbl}
\usepackage{makecell}
\usepackage{arydshln}
\usepackage{subfigure} 
\usepackage{booktabs}
\usepackage{amsmath}
\usepackage{algorithm}  
\usepackage[switch]{lineno} 
\usepackage[noend]{algpseudocode}
\usepackage{varwidth}
\usepackage{multirow}
\usepackage{amsfonts}

\newcommand{\tabincell}[2]{\begin{tabular}{@{}#1@{}}#2\end{tabular}}
\usepackage{ragged2e}

\begin{document}
%
\title{Multi-turn Dialogue Comprehension \\from a Topic-aware Perspective}
\author{Xinbei Ma, Yi~Xu, Hai~Zhao$^*$, Zhuosheng Zhang
\IEEEcompsocitemizethanks{\IEEEcompsocthanksitem Xinbei Ma, Yi Xu, Hai Zhao, and Zhuosheng Zhang are with the Department of Computer Science and Engineering, Shanghai Jiao Tong University, and also with Key Laboratory of Shanghai Education Commission for Intelligent Interaction and Cognitive Engineering, Shanghai Jiao Tong University, and also with MoE Key Lab of Artificial Intelligence, AI Institute, Shanghai Jiao Tong University. E-mail: sjtumaxb@sjtu.edu.cn, xuyi\_2019@sjtu.edu.cn, zhaohai@cs.sjtu.edu.cn, zhangzs@sjtu.edu.cn. ($^*$Corresponding author: Hai Zhao.)
        \IEEEcompsocthanksitem  This paper was partially supported by Key Projects of National Natural Science Foundation of China (U1836222 and 61733011).
        	 \IEEEcompsocthanksitem Part of this study has been accepted as "Topic-Aware Multi-turn Dialogue Modeling" \cite{xu2021topic} in the Thirty-Fifth AAAI Conference on Artificial Intelligence (AAAI 2021). This paper moves to a higher level of comprehending and modeling multi-turn dialogues from a topic-aware perspective rather than keeping limited to the response selection task in \cite{xu2021topic}, where our topic-aware method only benefits a single dialogue. This paper further conducts topic-aware clustering on all dialogue data to help understand the topic focus, which has many applications in reality. 
}

}
\IEEEtitleabstractindextext{%
\begin{abstract}
\justifying 
Dialogue related Machine Reading Comprehension requires language models to effectively decoupling and modeling on multi-turn dialogue passages. As a dialogue development goes after the intentions of participants, its topic may not keep constant through the whole passage. Hence, it is non-trivial to detect and leverage the topic shift in dialogue modeling.
Topic modeling, although has been widely studied in plain text, deserves far more utilization in dialogue reading comprehension.
This paper proposes to model multi-turn dialogues from a topic-aware perspective. We start with a dialogue segmentation algorithm to split a dialogue passage into topic-concentrated fragments in an unsupervised way.
Then we use these fragments as topic-aware language processing units in further dialogue comprehension.
On one hand, the split segments indict specific topic rather than mixed intention, thus show convenient on in-domain topic detection and location. For this task, we design a clustering system with a self-training auto-encoder, and we build two constructed datasets for evaluation.
On the other hand, the split segments are appropriate element of multi-turn dialogue response selection. For this purpose, we further present a novel model, Topic-Aware Dual-Attention Matching (TADAM) Network, which takes topic segments as processing element and matches response candidates with a dual cross-attention. Empirical studies on three public benchmarks show great improvements over baselines.
Our work continues the previous studies on document topic, and brings the dialogue modeling to a novel topic-aware perspective with exhaustive experiments and analyses.
\end{abstract}

\begin{IEEEkeywords}
Multi-turn Dialogue Modeling, Topic-aware, Segmentation, Clustering, Response Selection.
\end{IEEEkeywords}}

\maketitle

\IEEEdisplaynontitleabstractindextext

%
\IEEEpeerreviewmaketitle

\IEEEraisesectionheading{\section{Introduction}\label{sec:introduction}}

\IEEEPARstart{M}{\textbf{otivation.}} People are always engaging in conversations in both real life and Internet space, resulting in massive multi-turn dialogue data. Dialogue, as the most convenient and effective communication method, is constructed by speakers with various intentions, such as consulting, discussion, reservation, etc. As a result, throughout a dialogue record, there are always multiple topics shifting along with the change of speaker intentions. Such topic shifts naturally happens after each several turns \cite{li-etal-2017-dailydialog}. We present an example in Table \ref{tab:dialogue_case}, which is a dialogue fragment between a customer and a shop assistant from E-commerce \cite{yuan2019multi}. They first bargain in Turn-1 to Turn-6 and then talk about accompanying tools in Turn-7 to Turn-9. In another word, the topic changes from \textit{price} to \textit{tool} as the intention of the customer changes.

\begin{table}
    \centering
        \caption{ \label{tab:dialogue_case}A Multi-turn dialogue from E-commerce in \cite{yuan2019multi}. The topic changes from "price" to "tool" after “Turn-6”.  
    }
    \begin{tabular}{cl}
    \toprule
           \textbf{Turns} & \multicolumn{1}{c}{\textbf{Dialogue Text}} \\
           \midrule
           Turn-1 & A: \textit{Are there any discounts activities recently?} \\
        Turn-2 & B: \textit{No. Our product have been cheaper than before.} \\
        Turn-3 & A: \textit{Oh.} \\
        Turn-4 & B: \textit{Hum!} \\
        Turn-5 & A: \textit{I'll buy these nuts. Can you sell me cheaper?} \\
        Turn-6 & B: \textit{You can get some coupons on the homepage. }\\
        \hdashline 
        
        Turn-7 & A: \textit{Will you give me some nut clips?} \\
        Turn-8 & B: \textit{Of course we will.} \\
        Turn-9 & A: \textit{How many clips will you give?} \\
        \arrayrulecolor{black} \bottomrule
    \end{tabular}
\end{table}

Following previous studies on dialogue system,
researches on dialogue related Machine Reading Comprehension (MRC) arise and draw increase interest \cite{zhang2021advances}. 
Wide variation of reading comprehension tasks have been proposed on multi-turn dialogues, including response selection \cite{lowe-etal-2015-ubuntu,wu-etal-2017-sequential,zhang-etal-2018-modeling}, conversation-based question answering \cite{sun2019dream,reddy2019coqa,choi-etal-2018-quac}, emotion detection \cite{} etc. 
However, the character of topic transition is not fully considered in dialogue MRC. 
Although topic representation has been well explored in terms of plain text or documents, there is still a gap for applying it to dialogue. 
First, dialogues are topic-mixed as topic changes following the speaker intention, which requires a transition detection before modeling the topic.
Second, the segmented dialogue fragments are short (compared to documents), causing words sparsity problem. Also, topic transition annotation and representation on dialogue dataset is much more resource-consuming than on documents, making supervised training limited by inadequate data. In this paper, we aim to model the topic character of dialogue, and then utilize it as an enhancement on MRC.

\noindent \textbf{Prior Work and Limitations.}
Previous study of topic modeling are mainly applied to document topic detection or auxiliary modeling for dialogue generation. 
Previous Topic Detection and Tracking (TDT) \cite{yu2007topic} task focuses on detecting hot topics and finding topically related material in a stream of data during a specific period. TDT methods usually highly depend on lexical features which thus are not applicable to dialogue segments with short length and much noise or irregularity \cite{choi-etal-2001-latent, galley2003discourse,li2010netnews, chen2007hot}.
In multi-turn dialogue modeling, however, a long enough multi-turn dialogue may have multiple topics as the conversation goes on, and topic shift naturally happens by all means. Even though there are existing works that model semantic-relevant information \cite{yuan2019multi} or that extract topic words for better matching \cite{wu2018response}, the local topic-aware features are all limited on word and utterance-level. The constraint on lacking annotated topic labels hinders explicit topic modeling, leaving it in the level of auxiliary task \cite{ZhuP0ZH20, XingWWLHZM17, Dziri}.
In this paper, we design an unsupervised topic detection method, and explicitly use the topic-aware segments as language unit in further comprehension.

\noindent \textbf{Method and Contribution.} We propose to comprehend and model multi-turn dialogues from a topic-aware perspective. Through detecting topic shift point, a dialogue can be segmented into a set of topic segments, which consist of several continuous utterances. Our topic shift detection method run in an unsupervised way, avoiding the problem of lack of topic labels. Then we address two MRC tasks by change the language unit to the segmented dialogue fragments.

On the one hand, we consider in-domain task-oriented dialogue, such as customer service system. The topic-mixed dialogue records may be long and confusing for understanding. But the segmented dialogue fragments are topic-concentrated and self-informative, provide suitable unit for use intention detection. For example, security consultants can get rid of long, topic-mixed conversations and easily get consulting topics with related dialogue segments.
We apply Smooth Inverse Frequency (SIF) \cite{arora2016simple} followed by an pre-trained autoencoder to represent topic segments as embeddings. Then we combine clustering algorithm with autoencoder fine-tuning perfrom a hot topic detection. For evaluation, two datasets in Chinese and English are respectively constructed. 

On the other hand, we consider chatting record comprehension. Topic shift detection is capable of tracking global topic flow throughout entire dialogue records at discourse-level, thus can be utilized to help multi-turn dialogue comprehension.
Response selection requires model to select a most proper response from a collection of candidate answers according to the multi-turn dialogue history, which is the most important task in retrieval-based dialogue systems 
\cite{zhou2016multi,wu-etal-2017-sequential,zhou2018multi,zhu-etal-2018-lingke,zhang-etal-2018-modeling,zhang2018gaokao,tao2019multi,gu2019interactive,zhang2020neural,zhang2019sgnet}.  
Early studies mainly match response candidates with concatenated context \cite{lowe-etal-2015-ubuntu,kadlec2015improved,10.1145/2911451.2911542,tan2015lstm,10.5555/3060832.3061030,wang-jiang-2016-learning}. 
And recent works turn to interplay candidates and utterances to improve the matching score \cite{zhou2016multi,wu-etal-2017-sequential,zhang-etal-2018-modeling,zhou2018emotional,zhou2018multi,tao2019multi,yuan2019multi}.
We hold a really different angle from previous work to explicitly extracts topic segments from the dialogue history as basic units for further matching. Accordingly we design a novel model, \textbf{T}opic-\textbf{A}ware \textbf{D}ual-\textbf{A}ttention \textbf{M}atching (TADAM). The model takes segmented dialogue fragments as input, and weights fragments by both word-level and segment-level relevance, then matches the response candidates through a dual cross-attention module.
For evaluation, TADAM is applied on three benchmark datasets, ubuntu \cite{lowe-etal-2015-ubuntu}, Douban \cite{wu-etal-2017-sequential}, and E-commerce \cite{zhang-etal-2018-modeling}. And our proposed topic-aware modeling method achieves significant and stable improvements over baselines as experimental results shows. 
To our best knowledge, this is the first attempt of handling multi-turn dialogues in a topic-aware perspective. And we prove its validation and significance on two realistic dialogue MRC scenarios, hot topic detection and response selection.
The contribution of this paper can be summarized as:
(1) an idea to change the language unit into topic-aware segments, with an unsupervised topic-aware segmentation algorithm for dialogue records; (2) a novel task based on topic segments, in-domain hot topic detection; (3) bilingual datasets annotated topic transition;
(4) a novel model, TADAM, implements response selection task on the level of topic-aware segments, and outperforms various baselines; (5) exhaustive analysis of each approach from topic perspective.

\section{Related work}
\subsection{Topic Modeling}
Previous works on topic modeling approaches are mainly aimed to extract \textit{topic words}. Mostly these methods weights a document to select words that can summarize the topic \cite{twitterldaZhao, DBLP:conf/acl/LauBC17}.
Existing topic-related segmentation methods on plain text vary in how they represent sentences and how they measure the lexical similarity between sentences \cite{joty2013topic}. TextTiling \cite{hearst1997texttiling} proposes pseudo-sentences and applies cosine-based lexical similarity on term frequency. Based on TextTiling, LCSeg \cite{galley2003discourse} introduces lexical chains \cite{morris1991lexical} to build vectors. To alleviate the data sparsity from term frequency vector representation, Choi et al. (2001) \cite{choi-etal-2001-latent} employ Latent Semantic Analysis (LSA) for representation and Song et al. (2016) \cite{Song2016DialogueSS} further use word embeddings to enhance TextTiling. These segmentation methods are designed for plain text and ignore the sequential and inconsistent nature of dialogue turns. 

An related task is Topic Detection and Tracking (TDT) \cite{yu2007topic}, which focuses on detecting hot topics and finding related materials in a stream of data during a specific period. TDT highly depends on lexical features and time information of plain text data stream to extract hot terms and story clusters \cite{choi-etal-2001-latent, galley2003discourse,li2010netnews,chen2007hot}, which is not applicable to relatively short dialogues of casual and various expressions, and without time information. 

With a different motivation, our work pays efforts on detecting the changes of topics to model the inherent discontinuity of dialogues. Thus we focus on the transition prediction rather than representation. Also the segmented topic-concentrated fragments are much more shorter than normal passages, which hinders the topic representation.
Due to the lack of topic-annotated dialogue datasets, unsupervised learning is an alternative.
Related text clustering methods begin with Bag-Of-Words (BOW), whose data sparsity problem \cite{hadifar-etal-2019-self} has been solved by well-designed neural word embedding \cite{xu2017self}.

\subsection{Machine Reading Comprehension on Dialogues}
Dialogues remains to be a challenge for Machine Reading Comprehension (MRC) as its flexibility, inconsistency and discontinuity. The speaker transition happens from time to time, leading the development of a dialogue record. Such special features are intractable for MRC tasks that are well-solved on plain text, and thus raise attention on dialogue comprehension scenario. 
Without loss of generality, we focus on response selection as a representative task of dialogue MRC, where a most proper response is required to be selected among all candidates.
Earlier studies conduct single-turn match \cite{lowe-etal-2015-ubuntu, kadlec2015improved}. All context utterances are concatenated like a plain text and then matched with the candidates \cite{10.1145/2911451.2911542,tan2015lstm,10.5555/3060832.3061030,wang-jiang-2016-learning}. 
More recent methods explore relationship among utterances sequentially, and gradually form a framework of \textit{Representation-Matching-Aggregation} \cite{zhou2016multi,wu-etal-2017-sequential,zhang-etal-2018-modeling,zhou2018emotional,zhou2018multi,tao2019multi,yuan2019multi}.  

\textit{Representation}, to get the language embedding, always falls in two main methods, to encode each utterance separately, or to encode the whole context using Pre-trained Language Models (PrLMs) \cite{devlin2018bert,liu2019roberta, clark2020electra}, and then split out each utterance \cite{zhang2020SemBERT,zhang2021retro}. 
\textit{Matching} module are mainly based on attention mechanism and broadly tried.
For example, 
DAM \cite{zhou2018multi} matches a response with its multi-turn context entirely based on attention. MSN \cite{yuan2019multi} filters out irrelevant context to reduce noise. 
And the \textit{Aggregation} layer fuses all features and derives the matching score.

Given the unique of dialogue passages, many works have achieved performance improvement by modeling specific characters such as speaker property \cite{Gu:2020:SABERT:3340531.3412330} and discourse parsing \cite{}. But they still stick to model dialogues on utterance-level and less attention is paid to topic transition. Differently, our work turns to segment-level from a topic-aware aspect, and proves the effectiveness of topic modeling.

\subsection{Topic Representation in Dialogues}
In studies of dialogue system, topic has been incorporated with chatting records to refine dialogue generation quality \cite{luan2016lstm,serban2017hierarchical, xingTA, sevegnani-etal-2021-otters,chengunrock}. And recent works on MRC follow topic-aware methods in tasks such as response selection \cite{wu2018response} and emotion detection \cite{ZhuP0ZH20}.
Most existing MRC works can be divided into two patterns. The first is word-level topic representation, mainly based on topic word models like Twitter LDA \cite{twitterldaZhao}. 
Many dialogue generation model such as R-LDA-CONV \cite{DBLP:journals/corr/LuanJO16}, TA-seq2seq \cite{XingWWLHZM17}, A-RNN \cite{DBLP:conf/aaai/MeiBW17} use LDA-based topic words or topic vectors to improve the topic concentration when generating continuation of conversations.
Wu et al. (2018) \cite{wu2018response} introduce topic vectors in the matching step, which are linear combinations of topic words from the context and the response respectively. 
A matching framework \cite{wu2018knowledge} is proposed that can inject topic words as external knowledge. 
The second is auxiliary topic embedding.
In VHRED \cite{serban2017hierarchical}, a Variable Auto-Encoder (VAE) is fine-tuned for topic representation learning, and it borrowed as auxiliary embedding in emotion detection \cite{ZhuP0ZH20}. A topic attention is proved beneficial for response selection and dialogue disentanglement \cite{wang2020response}.

Besides, existing explicit modeling methods for dialogue topics may has some issues. The topic classification on open-domain \cite{chengunrock} needs a pre-setting and is very coarse across each domains. The topic prediction in Wang et al. (2020) \cite{wang2020response} is close to a disentangled conversation selection and remains on word-level.

Different from all the previous studies, this work for the first time proposes a novel approach which explicitly extracts topic-aware segments as processing units and thus is capable of globally handling topic clues at discourse-level. Our topic-aware model accords with realistic dialogue scenes where topic shift is a common fact as a conversation goes on.

\section{Topic-aware Modeling Approach}
\subsection{Task Formulation}
Our topic-aware modeling approach begins with a segmentation algorithm for multi-turn dialogue, and then we use the segmented dialogue fragments as processing units in two MRC tasks, hot topic detection and response selection. Examples of each are presented in Table \ref{tab:tasks}.

\begin{table}[tbh]
    \centering
        \caption{ \label{tab:tasks}Examples for segmentation, hot topic detection, and response selection.
    }
    \begin{tabular}{ll}
    \toprule
       \multicolumn{2}{c}{{\textbf{Task \uppercase\expandafter{\romannumeral1}: Segmentation}}}\\
       \midrule
            \textbf{Input} & A dialogue. \\
           \midrule
        Turn-1 & A: \textit{I got it, it tastes good.} \\
        Turn-2 & B: \textit{Thanks.} \\
        \hdashline 
        Turn-3 & A: \textit{My friends asked me to buy some for them, can I} \\
         & \textit{\quad get cheaper if I buy more?}\\
        Turn-4 & B: \textit{Dear, this is the lowest price.} \\
        \hdashline 
        Turn-5 & A: \textit{I want to buy 5 more packs. Can we not use China}\\
        & \textit{ \quad  Post? It's too much trouble.}\\
        Turn-6 & B: \textit{You can skip the postal service. } \\
        Turn-7 & A: \textit{It requires a signature. This is too troublesome.} \\
        Turn-8 & B: \textit{Please place your order and I will make the note.}\\
        Turn-9 & A: \textit{OK don't forget.} \\
        \midrule
        \textbf{Prediction} & Turn-3, Turn-5.\\
        \midrule
        \multicolumn{2}{c}{{\textbf{Task \uppercase\expandafter{\romannumeral2}: Hot Topic Detection}}}\\
        \midrule
        \textbf{Input} & Segments cut from in-domain dialogues.\\
        \midrule
        Seg.-1 & \tabincell{l}{\textit{Is there rain on Sunday? What city would you like the }\\\textit{forecast for? Compton. It will not rain on Saturday in} \\\textit{Compton.}}\\
        \hdashline
        Seg.-2 & \tabincell{l}{\textit{When is my doctor's appointment? Your doctor's appoint-}\\ \textit{ment is Monday at 1 pm.}}\\
        \hdashline
        Seg.-3 & \tabincell{l}{\textit{Hello, do you know where the parkside police station is?} \\\textit{It is in Parkside, Cambridge, CB11JG Can I get the }\\\textit{phone number? Their contact number is 01223358966.}}\\
        \hdashline
        Seg.-4 & \tabincell{l}{\textit{I need the location of a local hospital. The Addenbrookes} \\\textit{Hospital is located at Hills Rd, Cambridge Postcode} \\\textit{ CB20QQ.}}\\
        ... & ...\\
        \midrule
\textbf{Prediction}&\tabincell{l}{(Seg.-1, weather), (Seg.-2, schedule), (Seg.-3, police),\\ (Seg.-4, hostipal), (..., ...)} \\
        \midrule
        \multicolumn{2}{c}{{\textbf{Task \uppercase\expandafter{\romannumeral3}: Response Selection}}}\\
        \midrule
        \textbf{Input} & A dialogue, response candidates. \\
           \midrule
        Turn-1 & \textit{hey .. i installed a pata card in my comp with \_number\_ ...} \\
        Turn-2 & \textit{thats a bit hard because its a different comp with ...} \\
        Turn-3 & \textit{it is sata or pata ??} \\
        Turn-4 & \textit{its a sata card with a pata connector as well.} \\
        ... & ...\\
        Turn-10 & \textit{nope modprobe pata\_via module pata\_via not found. } \\
        \hdashline
        Resp.-1 & \textit{that is already loaded.. part of ide\_core in lsmod...} \\
        Resp.-2 & \textit{or an option or something it's pretty obvious...} \\
        ... & ...\\
        Resp.-$r$ & \textit{if you are of a mind to use the package manager why not ...} \\
        
        \midrule
        \textbf{Prediction} &(Resp.-1, 1), (Resp.-2, 0), ..., (Resp.-1, 0)\\
        \arrayrulecolor{black} \bottomrule
    \end{tabular}
\end{table}

\noindent \textbf{$\bullet$ Segmentation.}
Segmentation is our fundamental task. Given a continuous multi-turn dialogue history $C$ with $n$ utterances, $C=\{u_1, ..., u_{n}\}$. As shown in Table \ref{tab:tasks}, we aim to detect the transition points (like the Turn-6) and cut the dialogues into more self-informative segments. Let $G= \{0, g_1, g_2, ..., g_{t-1}, n\}$ be the transition point set, and the segments can be denoted as $S = \{S_i\}= \{ u_{g_i-1}, ..., u_{g_{i}}\}$.

\noindent \textbf{$\bullet$ Hot Topic Detection.}
We build this task with the realistic consideration that in-domain conversations usually follows some certain purposes of speakers and leads to topics that are common and frequent, which we refer to \textit{hot topics}. Detecting and locating the hot topics may makes it more convenient to response, instead of scanning the whole chatting logs. Inheriting the segments $S$, our goal is to well encode each $S_i$ into $z_i$. Due to the heavy consumption of data annotation, we cluster ${z_i}$ into hot groups for evaluation.

\noindent \textbf{$\bullet$ Response Selection.}
As a representative task of MRC, response selection finds the most suitable response among candidates. We denote the dataset as $D = \{ {(C_k,r_{kj},y_{kj})}_{j=1}^c\} _{k = 1}^N$, where $C$ is dialogue context, and $r$ is the candidate response, and $y \in \{ 0,1\}$ is the label indicating whether $r$ is a best response for $C$. $N$ is the dialogue number and $c$ is the candidate number. And our goal is to predict the label $y$ .The processing units also follow the segmentation, $C = \{ {S_1},{\rm{ }}...,{S_t}\}$ and ${S_i},1 \le i \le t$ is the $i$-th topic segment in context $C$.

\subsection{Topic-aware Dialogue Segmentation}
\label{segintro}
\subsubsection{Segmentation Algorithm}
Our topic-aware dialogue segmentation algorithm greedily checks adjacent utterances to determine a segmentation that lets resulting segments mostly differ.
As is shown in Algorithm \ref{DTSG}, topic transition points are detected in an unsupervised way.
The segmentation algorithm is based on utterance $n$-grams. 
To avoid yielding too many fragments, we set the check interval as $k$ utterance, i.e. regard $k$ utterances as a piece. Then we check all cut points between each adjacent pieces in an iteration. For one current piece, we find the most possible piece as the end of this topic segment, where we calculate the similarity between this possible segment and it context on both sides. The context are a fixed-length $d$ of utterances on the left and right. Then we save such a piece as a transition point.

\begin{algorithm}[tbh]
\caption{Topic-aware Segmentation Algorithm}
\label{DTSG}
\begin{algorithmic}[1] 
\Require
Dialogue $C=\{{u_1},{u_2},...,{u_{n}}\}$
\Ensure 
Transition point list $G$, Segments list $S$
\State $G$=[ ], $S$=[ ], start index $i=1$
\While {$i \le  n $}
\State ${l} = {u_{i - d}} \oplus  \cdots  \oplus {u_{i- 1}}$
\State $j=1, {c_0}=``",$
\While {$i+j \le n+1$ \textbf{and} $j \le R$}
\State $c_j = {c_{j-1}} \oplus {u_{i+j-1}}$
\If {$j$ mod $k$ == 0}
\State $r_j = {u_{i + j }} \oplus  \cdots  \oplus {u_{i + j + d-1}}$
\State \begin{varwidth}[t]{\linewidth}
$cos{t_{{c_j}}} = \max($ 
${\rm sim}(E({c_j}),E(l))$, \\
${\rm sim}(E({c_j}),E({r_j})))$
\end{varwidth}
\EndIf
\State $j+=1$
\EndWhile
\State $j^ *  = \mathop {\min }\limits_{{j}} cos{t_{{c_j}}}, c_j^*  = c_{j^*}$
\State $G.{\rm append}(j^ *), i= j^* +1$
\State $S.{\rm append}(c_j^ *)$
\EndWhile
\Return $G, S$
\end{algorithmic}
\end{algorithm}
 
\begin{figure}
\centering
\includegraphics[width=0.48\textwidth]{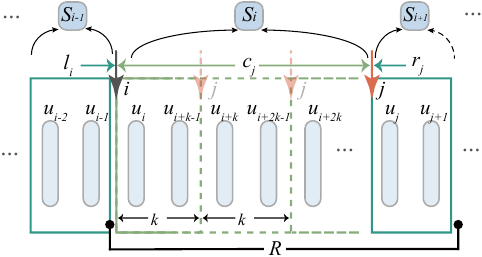}
\caption{A schematic diagram for the segmentation algorithm}
\label{fig:segment}
\end{figure}
Additional detailed settings are implemented as is illustrated in Figure \ref{fig:segment}.
$R$ is set to control the number of utterances in each topic segment and time consumption.
We also set a threshold $\theta$, so that if the minimum similarity is still bigger than the threshold, i.e. $cost_{min} > \theta$, then we skip this segmentation.
The encoding method of dialogue text is denoted as $E(\cdot)$ in Algorithm \ref{DTSG}, which we use word embedding (e.g. GloVe \cite{pennington2014glove}) or a PrLM (e.g. BERT \cite{devlin2018bert}).
As a result, a dialogue is cut into a set of segments as utterance $n$-grams, which are likely to hold different topics. We refer to such segments as topic(-aware) segment in the following paper.

\subsubsection{Assistance Training}
To provide a more topic-related encoder for $E(\cdot)$ in the algorithm \ref{DTSG}, we use a self-supervised training objective to enhance a PrLM. We concatenate sentences from two passages whose topics are different, and ask the language model to predict the topic transition point, i.e., the beginning sentence of the latter topic. Specifically, we use two random passages from Wiki corpus and extract sentences from them as input. Sentences are separated with a special token indicating sentence ends. And the position of the special token before the latter topic is regarded as the transition point. Using this self-supervised task, we further train a PrLM to work as an encoder.

\subsection{Hot Topic Detection}
\label{clustering_sec}
Continuing the topic-aware segmentation, we further build a encoding system for the topic-concentrated segments also in unsupervised learning. Hot topic detection is proposed as an evaluation task from a realistic point of view, and we verify our representation on it. More implementation detailed are introduced in \textit{Experiment and Analysis} section.

\begin{figure*}
\centering
\includegraphics[scale=1]{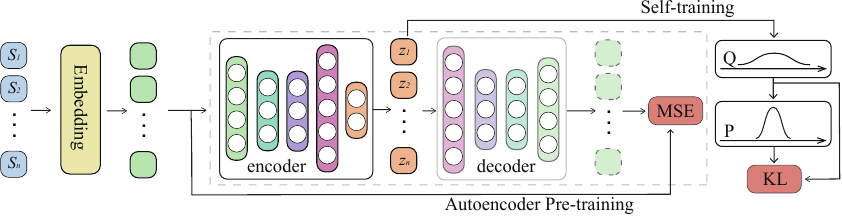}
\caption{Framework of segmentation encoding and hot topic detection.
}
\label{cluster_figure}
\end{figure*}

Framework of this section consists of two parts shown in Figure \ref{cluster_figure}:
(1) Embedding layer: Smooth Inverse Frequency (SIF) algorithm segments encoding. 
(2) Autoencoder (AE): An encoder-decoder architecture that is pre-trained for further representation learning. 
With the goal of hot topic clustering, the segments ${S_i}$ are input to the framework above. 
First we pre-train the encoder and decoder of AE, where the encoder learns a projection from input space to a latent space $Z$, and the decoder learns reconstruction.
And the output ${z_i}$ of the AE encoder are used as representations in $k$-means. During the convergence of $k$-means, we jointly fine-tune the encoder in a self-training way.
This part follows existing works on short text clustering \cite{hadifar-etal-2019-self, xie2016unsupervised}.
\subsubsection{Model Framework}
\label{BERT-SIF sec} 
\textbf{$\bullet$ Embedding Layer.}
Inheriting the segmentation, our linguistic processing unit is the topic-aware segments ${S_i}$. And each segment consists of continuous utterances, which are concatenated for encoding. We apply Smooth Inverse Frequency algorithm (SIF) \cite{arora2016simple} as shown in Algorithm \ref{SEC}. SIF instead of TF-IDF is used to re-weigh embeddings , where $a$ is the smooth parameter and $f(\cdot)$ is word frequency. $E(\cdot)$ denotes the word embedding (e.g. PrLM \cite{devlin2018bert} or GloVe \cite{pennington2014glove}). 
The final text embedding is obtained after subtracting the projection to their first principal component. This operation removes common information of the embeddings and reserves the difference, so that a topic segment differs more, compared to those of other topics.

\begin{algorithm}[bht]
\caption{SIF Representation}
\label{SEC}
\begin{algorithmic}[1] 
\Require 
A set of short texts $\mathbb{T}$, word frequency $\left\{ {f(w):w \in V} \right\}$
\Ensure
Text embeddings $\left\{ {{v_t}:t \in \mathbb{T}} \right\}$
\For {\textbf{each} text $t$ in $\mathbb{T}$}
\State $v_w$=$E(w)$ for $w \in t$
\State ${v_t} = \frac{1}{{\left| t \right|}}\sum\nolimits_{w \in t} {\frac{a}{{a + f(w)}}} {v_w}$
\EndFor
\State Form matrix  with $\left\{ {{v_t}:t \in \mathbb{T}} \right\}$ as columns and get its first singular vector $\hat{v}$
\For {\textbf{each} text $t$ in $\mathbb{T}$}
\State ${v_t}={v_t} - \hat{v}{\hat{v}^T}{v_t}$
\EndFor
\end{algorithmic}
\end{algorithm}
\noindent \textbf{$\bullet$ Autoencoder.}
Following Deep Embedded Clustering (DEC) \cite{hadifar-etal-2019-self}, stacked autoencoder (SAE) is applied as the autoencoder module. The encoder and decoder of an SAE is a stack of fully connected layers with \texttt{ReLU} activation function. The function of SAE lies in two aspects: (1) automatically learn a projection to the latent feature space $Z$ as better embedding; (2) reduce the dimension of embedding for further processing.

\subsubsection{Model Training}

\textbf{$\bullet$ Pre-training.}
First the SAE is pre-trained.
As is known, the encoder part learns to transform SIF input to a latent feature space $Z$, while the decoder maps the lower dimensional vectors into original space to reconstruct the input vectors. The training objective is to minimize the reconstruction loss between the decoder output and the input. Here we use least-squares loss.

\noindent \textbf{$\bullet$ Self-training.}
At this stage, the vectors ${z_i}$ are not directly clustered.
Instead, we expect ${z_i}$ to be self-refined by further fine-tuning the autoencoder to produce more distinguishable embeddings for clustering. 
Thus we adopt a self-training method \cite{xie2016unsupervised} that is first proposed for image processing. The self-training process is concluded in Algorithm \ref{self-train-al}.

We first conduct  $k$-means clustering algorithm to find $m$ cluster centroids ${\mu_j}$.
For each centroid $\mu_j$, 
we use the Student’s $t$-distribution  
to calculate similarity of each segment embedding $z_i$, where  $z_i$ is the output of the encoder part:
Then we measure the possible that each segment ($z_i$) is clustered into each group ($j$-th group) by calculating the Student’s $t$-distribution. 
This distribution stands for the results of clustering, and we denote it as distribution $Q$. The formulation is
\begin{equation}
\label{q}
{q_{ij}} = \frac{{{{(1 + {{{{\left\| {{z_i} - {\mu _j}} \right\|}^2}} \mathord{\left/
 {\vphantom {{{{\left\| {{z_i} - {\mu _j}} \right\|}^2}} \alpha }} \right.
 \kern-\nulldelimiterspace} \alpha })}^{ - \frac{{\alpha  + 1}}{2}}}}}{{\sum\nolimits_{j'} {{{(1 + {{{{\left\| {{z_i} - {\mu _{j'}}} \right\|}^2}} \mathord{\left/
 {\vphantom {{{{\left\| {{z_i} - {\mu _{j'}}} \right\|}^2}} \alpha }} \right.
 \kern-\nulldelimiterspace} \alpha })}^{ - \frac{{\alpha  + 1}}{2}}}} }}
\end{equation}
, where $\alpha$ is the degree of freedom of the Student’s $t$-distribution.
Then we create a pseudo target distribution for the unsupervised training. We use the quadratic normalization of $Q$, and denote it as distribution $P$.
According to Xie et al. (2016) \cite{xie2016unsupervised}, $P$ puts more emphasis on data that are assigned with high confidence, and prevent large clusters from distorting the hidden feature space. 
$P$ can be formulated as
\begin{equation}
\label{p}
{p_{ij}} = \frac{{{{q_{ij}^2} \mathord{\left/
 {\vphantom {{q_{ij}^2} {\sum\nolimits_{i'} {{q_{i'j}}} }}} \right.
 \kern-\nulldelimiterspace} {\sum\nolimits_{i'} {{q_{i'j}}} }}}}{{\sum\nolimits_{j'} {({{q_{ij'}^2} \mathord{\left/
 {\vphantom {{q_{ij'}^2} {\sum\nolimits_{i'} {{q_{i'j'}})} }}} \right.
 \kern-\nulldelimiterspace} {\sum\nolimits_{i'} {{q_{i'j'}})} }}} }}
\end{equation}

Then we minimize KL divergence of distributions $P$ and $Q$ as training objection to fine-tune the autoencoder, pushing $Q$ to approach $P$.
\begin{equation}
\label{KL}
KL(P\left\| Q \right.) = \sum\limits_i {\sum\limits_j {{p_{ij}}\log \frac{{{p_{ij}}}}{{{q_{ij}}}}} } 
\end{equation}
Each iteration of the self-training gives a new segment embedding ${z_i}$ and thus a new cluster result. The training ends when clusters get stable, which are regard as the final group results.
\begin{algorithm}[h!]
\caption{Self-training Process}
\label{self-train-al}
\begin{algorithmic}[1] 
\Require
Segmentation vectors $\{ {z_i},1 \le i \le n\} $ from pre-trained AE; $m$ centroids $\{ {\mu _j},1 \le j \le m\}$ from $k$-means
\Ensure
ClusterID($i$) for each segment $i$
\State Get distribution $P$ w.r.t. Eq.(\ref{p})
\While {$iter$  \textless $iter_{max}$}
\State 
\begin{varwidth}[t]{\linewidth}
Get distribution $Q$ w.r.t. Eq.(\ref{q}) \\ 
Train AE w.r.t. Eq.(\ref{KL})
\end{varwidth}
\If {$iter$ mod 500 == 0 }
\State Update distribution $P$ and $Q$
\State 
\begin{varwidth}[t]{\linewidth}
For each segment $i$, \par
ClusterID($i$)= arg max$_j$ $q_{ij}$
\end{varwidth}
\If {No ClusterID update}
\State Break
\EndIf
\EndIf
\State $iter$ = $iter$ + 1
\EndWhile
\State Collect 
all ClusterID($i$) as clustering results.
\end{algorithmic}
\end{algorithm}

\subsection{Response Selection}
\begin{figure*}[ht]
\centering
\includegraphics[scale=1.0]{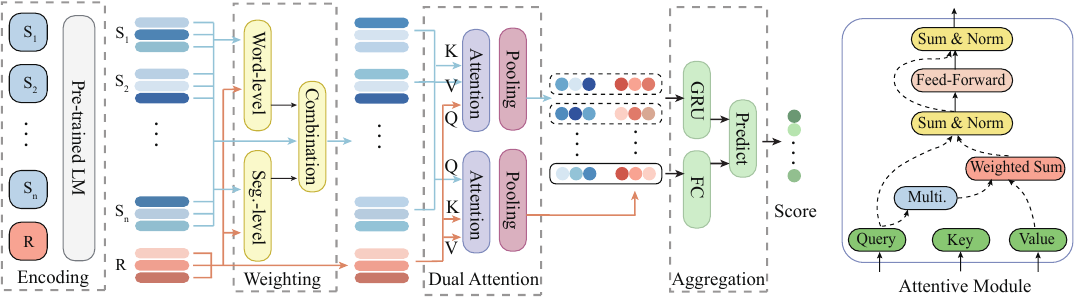}
\caption{Architecture of our TADAM network (left) and Attentive Module (right).}
\label{fig:Module}
\end{figure*}

Following the results of segmentation, we further use the self-informative topic-aware segments to address response selection problem. Based on the fact that topic shift at discourse level helps reading comprehension, we use the segments as language processing units, $C = \{ {S_1},{\rm{ }}...,{S_t}\}$. 
We propose \textbf{T}opic-\textbf{A}ware \textbf{D}ual-\textbf{A}ttention \textbf{M}atching (TADAM) network, where we enhance the attention between segments and response candidates.

Figure \ref{fig:Module} shows the overview of TADAM architecture, which follows the \textit{Representation-Matching-Aggregation} paradigm. 
In \textit{representation} layer, all segments and responses are concatenated as an input sequence of a PrLM encoder, and then separated back to segments according to position information. 
The \textit{matching} module includes two parts. 
(1) We first use responses to weight each segment at both word and segment levels, in order to pay draw attention to more related segments. 
(2) Then we use a dual cross-attention for deep matching of segments and responses.
In \textit{aggregation} module, 
a GRU is used for the fusion of all matching vectors, where the last segment is especially highlighted. 
Ultimately, final fused vector is passed to a linear layer for a matching score.
\subsubsection{Encoding and Separation}
We take well pre-trained contextualized language model as our encoder component. Following the encoding manner of PrLMs such as BERT \cite{devlin2018bert} and ALBERT \cite{Lan2020ALBERT:}, we concatenate all segments $\{ {S_{i}}\} _{i=1}^t$ and response $r$ with special tokens: $X = \left[ \texttt{CLS} \right]{S_1}\left[ \texttt{SEP} \right].{\rm{ }}.{\rm{ }}.\left[ \texttt{SEP} \right]{S_t}\left[ \texttt{SEP} \right]r\left[ \texttt{SEP} \right]$, which is fed into the encoder. 
Then we split the word vectors back into segments and response for further segment-level matching. 
As the number of segments $t$ is different among all contexts, and the length of each segment varies, we set the maximum number of segments in each context as $T$, and  the maximum length of segments as $L$. 
After encoding and separation, we get context representation $C_1 \in {\mathbb{R}^{T \times L \times d}}$ or $\{ {S_i} \in {\mathbb{R}^{L \times d}}\} _{i = 1}^T$, and response representation $r_1 \in {\mathbb{R}^{L \times d}}$, where $d$ is hidden dimension of the PrLM encoder. 

\subsubsection{Segment Weighting}
\label{sec: Segment Weight}
As the first part of the matching network,
the weighting module is proposed to improve the hard selection of utterances \cite{yuan2019multi}. Their work reserves relevant and meaningful utterances of context by comparing to a threshold, and deletes all other utterances.
Considering our process unit is segments, which are much longer than single utterances, the hard selection will bring information scarcity.
Instead, we assign a weight to each segments, using the candidate response as a key utterance on both word and segment granularities.

\noindent \textbf{$\bullet$ Word-level Weighting.}
At the word level, we build a matching feature map between each segment $S_i$ and response $r$, which is formulated as:
\begin{align}
\label{map}& {M} =\frac{1}{{\sqrt d }}{\rm tanh} [{\rm T}(C_1,W_1,r)] \times V_1, \\
\label{einsum}& {{\rm T}(C_1,W_1,r)_{xyuv}} = \sum\limits_k {{C_{1_{xyk}}} \times {W_{1_{kkv}}} \times {r_{uk}}}, \\
& {M_{pool}} = [ {\max M}_{\dim  = 2} , {\max M}_{\dim  = 3} ],
\end{align}
where equation \ref{einsum} is the Einstein notation of function $\rm T$ of equation \ref{map}. $W_1 \in \mathbb{R}^{d \times d \times h}, V_1 \in \mathbb{R}^h$ are learnable parameters.
The matrix $M$ is max-pooled in both row and column and then we get the matching map  ${M_{pool}} \in \mathbb{R}^{T \times 2L}$. 
$M_{pool}$ indicates the relevance between response $r$ and $T$ segments at the word level. Then we transfer matching features $M_{pool}$ into weights for $T$ segments through a linear layer:
\begin{equation}
{w_w} = {\rm softmax} (M_{pool}W' + b),
\end{equation}
where $w_w \in \mathbb{R}^{T}$ is segment weights at word level, and $W' \in \mathbb{R}^{2L},b \in \mathbb{R}^T$ are learnable parameters.

\noindent \textbf{$\bullet$ Segment-level Weighting.}
At the segment level, we build segment representation $C' \in \mathbb{R}^{T\times d}$ by mean-pool the token vectors, and calculate the cosine similarity with the response $r$ to get weights $w_s \in \mathbb{R}^{T}$ of segment level. $w_s$ catches the overall semantic similarity between the responses and segments.
\begin{align}
&C'={\rm mean}{(C_1)_{\dim  = 2}}, \\
&{w_s} = \cos (C',r),
\end{align}

\noindent \textbf{$\bullet$ Combination.}
Weights on word and segment level are complementary to each other, thus are sum up with a hyper-parameter $\beta$.
Then we multiply the sum $s$ and $C^1$ to get weighted segments representation that is consistent with the degree of relevance to the response. The weighted context are referred to as $C_2$.
\begin{align}
\label{beta}
&s=\beta w_w+(1-\beta)w_s, \\
&{C_2}=s \odot C_1,
\end{align}

\subsubsection{Dual Cross-attention Matching}
Our second part of the matching network is a dual cross-attention.
We are inspired by Attentive Module in DAM \cite{zhou2018multi}, which borrows transformer blocks \cite{vaswani2017attention} to model the interaction between two sequences. 
Here cross-attention method is applied to the segments and response in a dual way.

\noindent \textbf{$\bullet$ Attentive Module.}
The architecture of attentive module is shown in Figure \ref{fig:Module}, which takes three sequences as input: query sequence $Q\in \mathbb{R}^{n_q\times d}$, key sequence $K\in \mathbb{R}^{n_k\times d}$ and value sequence $V\in \mathbb{R}^{n_v\times d}$.
$n_q,n_k,n_v$ are the number of tokens respectively, and $d$ is hidden dimension.

Attentive module first takes each word in the query sentence to attend to words in the key sentence via Scaled Dot-Product Attention \cite{vaswani2017attention}, then weights attention scores upon the value sentence.
The attended value $V_{att} \in \mathbb{R}^{n_q\times d}$ is then passed to layer normalization \cite{ba2016layer}, whose output is denoted as $V'_{att} \in \mathbb{R}^{n_q\times d}$.
Then $V'_{att}$ is then fed into a Feed-Forward Network (FFN) with ReLU \cite{lecun2015deep} activation.
We denote the attentive module as $A$:
 \begin{align}
 & {A}(Q,K,V) = FFN(V'_{att}),\\
 & {V'_{att}} = Norm {(V_{att})}, \\
 & {V_{att}} = {\rm softmax} (\frac{{Q{K^T}}}{{\sqrt d }})V.
 \end{align}

\noindent \textbf{$\bullet$ Dual Attention.}
We apply symmetrical input to two attentive modules. One takes segments as $Q$, and the response candidate as $K$ and $V$, while the other takes the response candidate as $Q$, and segments as $K$ and $V$.
Note that our units of attention is segments.
In formula, the weighted context representation $C_2$ is split into segments ${S_{2_i}} \in \mathbb{R}^{L \times d}$ and perform dual attention with the response $r$. 
\begin{align}
&\{ S_{2_i}\}_{i = 1}^T = C_2,\\
&\tilde{S_i}={A}({S_{2_i}},r,r),\\
&\tilde{r_i}={A}(r,{S_{2_i}},{S_{2_i}}).
\end{align}
${\tilde S_i}$ and $\tilde r_i $ are attended representations and both in $\mathbb{R}^{L \times d}$.
Then we reconstruct the context-response pairs and mean-pool them to $C_3 \in \mathbb{R}^{T  \times 2d} $.
\begin{align}
\label{euq:C_S_1} &C_{3_s}=\{\tilde S_i\}_{i = 1}^T, \\
\label{euq:C_R_1} &C_{3_r}= \{ \tilde r_i\}_{i = 1}^T,\\
\label{euq:C}&{C_{3}} = [{{ \rm mean}(C_{3_s})}_{\dim  = 2} ,{{\rm mean}(C_{3_r})}_{\dim  = 2} ],
\end{align}
\subsubsection{Aggregation}
\label{sec: Aggregation}
In aggregation module, we use a GRU to model the relation of segments \cite{wu-etal-2017-sequential}. 
We also enhance the last segments and the response in aggregation. From the point of view that the last segment is nearest to a response in time sequence, it may be have closest relations. So the last ($T$-th) result of dual attention, $C_{3_T}$ is injected by a linear layer and concatenation.
\begin{align}
&\hat H = {\rm GRU}({C_{3}}),  \\ 
\label{euq:C_T}&C_{3_T} = [ {{\rm mean}( \tilde S_T)}_{\dim  = 1} , {{\rm mean}(\tilde R_T)}_{\dim  = 1} ],
\\
&\hat C_{3_T}=W_3C_{3_T}+b_3,
\end{align}
Then we get passage-level representation $\hat H \in \mathbb{R}^{2d}$.  $W_3 \in  \mathbb{R}^{2d \times 2d}, b_3 \in \mathbb{R}^{2d}$ are learnable parameters. 
Final score for context $C$ and its candidate response $r$ is predicted by a linear layer, where $W_4 \in  \mathbb{R}^{4d}, b_4 \in \mathbb{R}$ are parameters.
\begin{equation}
\label{euq:score}score = {\rm sigmoid}({W^T_4}[\hat H,\hat C_{3_T}] + {b_4}).
\end{equation}
Our training objective is a to minimize binary cross entropy loss between the $score$ and label $y$.

\section{Experiment and Analysis}
This section presents implementation of experiments as well as multi-angle analysis. This paper proposes a topic-aware segmentation method and accordingly gives novel solutions of two MRC tasks. Hence we introduce experiments by task: Section \ref{Seglab} for segmentation, Section \ref{clusterlab} for hot topic detection, and Section \ref{rslab} for response selection.
\begin{table}
\renewcommand\tabcolsep{14pt}
\centering
\caption{\label{seginfo} Statistics of topic transition datasets.}
\begin{tabular}{lll}
\hline 
\textbf{Statistics}&Chinese&English \\ 
\hline 
\#Dialogues & 505&711 \\
\#Utterances & 12867&19350 \\
\#Segments &2019&3465 \\
Avg. Tokens/Utter.  & 19.56&12.43\\
Avg. Utterance/Segment & 6.37&5.58 \\
Avg. Segment/Dialogue & 4.0&4.87\\
Avg. Utter./Dial.  & 25.48&27.22 \\
\hdashline
\#Topics & 13&10 \\
\hline
\end{tabular}
\end{table}
\begin{table}
\renewcommand\tabcolsep{2pt}
\centering
\caption{\label{topicinfo} Topics in our datasets for topic-aware clustering.}
\begin{tabular}{ll}
\hline 
\textbf{Dataset}&(Topic, Segment number) \\ 
\hline 
Chinese& \tabincell{l}{(\textit{Greetings},296), (\textit{Return visit},328), (\textit{Software operation},274), \\(\textit{Account setting\&deleting},120), (\textit{Identity check},268),\\ (\textit{Transaction advisory},244), (\textit{Sales department},41), (\textit{Stock},166), \\ (\textit{Urgent feedback},84), (\textit{Information modification},32), \\(\textit{Account problem},66), (\textit{Bankcard binding},40), (\textit{Rubbish},60)}\\
\midrule
English&\tabincell{l}{(\textit{navigate},400), (\textit{police},245), (\textit{hotel},438), (\textit{train},345), \\(\textit{schedule},414), (\textit{weather},398), (\textit{restaurant},442), \\(\textit{taxi},435), (\textit{hospital},287), (\textit{attraction},150)} \\
\hline
\end{tabular}
\end{table}

\begin{table*}[h!]
\centering
\caption{\label{cluster results} Dialogue hot topic clustering results}
\begin{tabular}{rllllllllllll}
\hline 
~&\multicolumn{2}{c}{\textbf{Model}} &~&\multicolumn{4}{c}{Chinese}&~&\multicolumn{4}{c}{English} \\ 
\cline{2-3}
\cline{5-8}
\cline{10-13}
 &Encoding&Clustering&~& \textbf{${N_c}$}& \textbf{$C_{rate}$} & \textbf{$A_{rate}$}& \textbf{$NMI$}&~& \textbf{${N_c}$}& \textbf{$C_{rate}$} & \textbf{$A_{rate}$} & \textbf{$NMI$}\\ 
\midrule
1&LDA&&& 3&\textbf{96.0}&35.1&32.0&~&5&95.1&52.7&77.3 \\
2&TF-IDF&$k$-means&& 4&41.9&22.2&19.0&~&9&93.4&69.1&65.1\\
3&GloVe&$k$-means&&6&60.0&27.4&24.4&~&8&90.6&54.7&64.7\\
4&BERT&$k$-means&&5&62.5&32.7&29.7&~&8&92.1&79.9&89.0\\
\midrule

5&GloVe+SIF&$k$-means&& 6&69.4&33.3&28.0&~&8&88.7&61.8&66.3\\
6&BERT+SIF&$k$-means&& \textbf{7}&72.0&38.9&33.4&~&9&97.9&83.1&84.4\\

7&GloVe+SAE&$k$-means&&6&57.5&29.7&29.0&~&8&92.9&66.2&73.1\\
8&BERT+SAE&$k$-means&&5&52.5&29.9&31.2&~&8&92.1&79.4&87.0\\
9&GloVe+SIF+SAE&$k$-means&& 6&61.4&26.6&28.9&~&8&91.3&70.6&78.6\\

10&BERT+SIF+SAE&$k$-means&& 6&53.5&35.7&33.8&~&9&95.5&80.6&86.7\\
\midrule
\midrule
11&GloVe&SAE+Self-training&&6&54.3&33.2&30.5&~&9&95.3&88.2&89.0\\
12&GloVe+SIF&SAE+Self-training&& \textbf{7}&70.6&36.7&33.5&~&9&91.6&74.4&77.5\\
13&BERT&SAE+Self-training&&6&62.5&37.2&33.4&~&8&90.9&78.4&87.9\\
14&BERT+SIF&SAE+Self-training&& 6&64.7&\textbf{42.5}&\textbf{36.6}&~&\textbf{10}&\textbf{100.0}&\textbf{91.9}&\textbf{92.0}\\
\hline
\end{tabular}
\end{table*}

\subsection{Experiment \uppercase\expandafter{\romannumeral1}: Segmentation}
\label{Seglab}
Experiments in this section evaluate the proposed topic-aware segmentation approach in Section \ref{segintro}, as a foundation of following two tasks.
\subsubsection{Dataset}
\label{dataconstruct}
Although our segmentation algorithm is unsupervised, we need labeled dataset to verify the effectiveness. The topic-aware segmentation is a novel task on dialogue and lack of off-the-shelf data. To fill this gap, we build topic transition datasets in both English and Chinese by our own. Our dataset includes 1.2k dialogues and statistics are shown in Table \ref{seginfo}.

(1) \textbf{Chinese Dataset}: Our Chinese dataset is derived from phone records of customer service on banking consultation. 
We manually annotate topic transition points of each dialogue. On average, each dialogue has 4 segments (3 transition points), and each segment includes 6.37 utterances. 
(2) \textbf{English Dataset}: The English topic transition dataset is constructed by joining existing dialogues. We concatenate single-topic dialogues that have no topic changes, mostly sticks on one single topic.
Specifically, we find single-topic dialogues from MultiWOZ Corpus\footnote{\url{https://doi.org/10.17863/CAM.41572}} \cite{budzianowski-etal-2018-multiwoz} and Stanford Dialog Dataset \cite{eric-etal-2017-key}. Note that we exclude multi-topic dialogue records from MultiWOZ. And the topic transition points are labeled as the concatenated points.
Redundant utterances like "\textit{thanks}" or "\textit{bye-bye}" are removed as we prefer dialogues to have more obvious topics.

\subsubsection{Settings}
\textbf{$\bullet$ Metrics.} 
For topic-aware segmentation, We adopt three Metrics:
(1) $MAE$ \cite{ijcai2018-612}  is defined as $\frac{1}{{\left| T \right|}}\sum\nolimits_{D \in T} {\left| {{N_{pred}}\left( D \right) - {N_{ref}}\left( D \right)} \right|} $, where $D$ is a dialogue, and ${N_{pred}}\left( D \right),{N_{ref}}\left( D \right)$ denote the prediction and reference number of segments $T$. 
(2) WindowDiff ($WD$) \cite{pevzner2002critique} uses a short window through the dialogue from beginning to end, and if the number of segmentation in prediction and reference is not identical, a penalty of 1 is added. 
The window size in our experiments is set to 4, and we report the mean $WD$ for all experiments. 
(3) $F_1$ score is the harmonic average of recall and precision of segmentation points.

\noindent \textbf{$\bullet$ Baselines.} 
We implement the segmentation using different encoding approach, word embedding of GloVe and BERT. But the algorithm requires sentence embedding. So for GloVe we simply use an average of word embeddings. For BERT, we try the mean-pooled vectors and the \texttt{[CLS]} vector. 
As our work is the first attempt on topic segmentation, we only compare to a previous method, TextTiling \cite{hearst1997texttiling}, which is a classic text segmentation algorithm using term frequency vectors to represent text. The three sentence embedding are combined with the two segmentation on both Chinese and English data, as is shown in Table \ref{tab: seg_results}.

\noindent \textbf{$\bullet$ Hyper-Parameters.} 
The BERTs we use for encoding are of base size BERT\footnote{https://github.com/huggingface/transformers} (bert-base-uncased \& bert-base-chinese). 
In both datasets, we set range $R=8$, jump step $k=2$, window size $d=2$ and the threshold $\theta= 0.6$. 
For TextTiling, the length of a pseudo sentence is set to 20 in Chinese dataset and 10 in English dataset, which is close to the mean length of utterances in both datasets. 
Window size and block size for TextTiling \cite{hearst1997texttiling} are all set to 6.

\subsubsection{Results and Analysis}
Results are shown in Table \ref{tab: seg_results}. We observe that: 
(1) Well pre-trained BERT works better than GloVe embedding. And \texttt{[CLS]} vector is more suitable for Chinese data while mean-pooled vector agrees with English data more.
The assistant training of PrLM (BERT$_{pt}$) contributes an improvement.
(2) In most cases, our segmentation algorithm surpasses TextTiling on all metrics.
(3) TextTiling tends to have a much larger $MAE$, because it ignores the number of turns in a topic round. 
TextTiling tends to have larger $MAE$, because it ignores the number of turns in a topic round. 
In the following topic-aware clustering task, 
we use algo.+BERT$_{CLS}$ for Chinese dataset and Our algo.+BERT$_{mean}$ for English dataset on hot topic clustering, and algo.+BERT$_{CLS}$ for all dataset on response selection for convenience.

\begin{table}[!ht]
  \renewcommand\tabcolsep{3.4pt}
            \centering
            \caption{\label{tab: seg_results} Topic-aware segmentation results.}
\begin{tabular}{lccccccc}
\toprule 
\multirow{2}*{\textbf{Method}}& \multicolumn{3}{c}{Chinese}&~&\multicolumn{3}{c}{English} \\ 
\cmidrule(lr){2-4}
\cmidrule(lr){6-8}
 ~& \textbf{$MAE$}& \textbf{~$WD$} & \textbf{~$F_1$}&~& \textbf{$MAE$}& \textbf{~$WD$} & \textbf{~$F_1$} \\ 
\midrule
TextTiling & 1.90&0.45  & 0.52&~&10.08& 0.83&0.34 \\
TextTiling+GloVe& 2.0&0.45  & 0.52&~&6.38& 0.75&0.33\\

TextTiling+BERT$_{mean}$& 6.50&0.60 & 0.45&~&9.64& 0.81&0.32\\
TextTiling+BERT$_{CLS}$& 6.51&0.60 & 0.45&~&9.78& 0.82&0.33\\
Our algo.+ GloVe&3.83&0.61&0.48&~&3.48&0.59&0.56\\
Our algo.+BERT$_{mean}$&2.95&0.52&0.51&~&2.98&\textbf{0.52}&\textbf{0.61}\\
Our algo.+BERT$_{CLS}$ &0.79 &0.34 &\textbf{0.61}&~ & \textbf{1.04}&0.54&0.44 \\
Our algo.+BERT$_{pt}$ &\textbf{ 0.74} &\textbf{0.34} &0.60&~ &  -- & -- & -- \\
\bottomrule
\end{tabular}
\end{table}
   
\subsection{Experiment \uppercase\expandafter{\romannumeral2}: Hot Topic Clustering}
\label{clusterlab}
In this experiment, we first evaluate the encoding of segments ${z_i}$ and topic clustering in Section \ref{clustering_sec}. Then we topic segments generated from our segmentation algorithm for end-to-end implementation.

\subsubsection{Dataset}
We use our topic transition datasets \ref{dataconstruct} and further annotate the topic labels of each transition point for clustering. 
There are respectively 13 and 10 topics in the Chinese and English dataset and each topic includes a fair number of segments. Table \ref{seginfo} and \ref{topicinfo} gives an overview.
The Chinese dataset is extracted from in-domain consultation records, thus can be labeled as hot topics such as \textit{Return visit}, \textit{Software operation}, etc. 
In addition, segments whose topic is too vague are excluded in clustering evaluation. 
Dialogues in English dataset are joint from passages with single and known topic. So we simply label their topic for segments.
\subsubsection{Settings}
\textbf{$\bullet$ Metrics.}
We use Hungarian algorithm  \cite{wong1983combinatorial} to map the nominal ClusterID returned by our topic clustering method to the true topics. Then we calculate the $F_1$ of each result cluster and set a threshold for $F_1$ (0.25 in our setting) to filter ones of low quality. The remaining groups are expected to have high rate of coverage for segments which are correctly clustered. For clustering, we adopt four metrics: (1) Number of remaining clusters, ${N_c}$. It means the number of topics we can cover. 
(2) Coverage rate of ${N_c}$ topic clusters, $C_{rate}$. 
(3) Coverage rate of segments which are accurately clustered in ${N_c}$ topic clusters, $A_{rate}$. $A_{rate}$ is of great importance and reflects the correctness of the clustering.
(4) Normalized Mutual Information, $NMI$. $NMI$ is to measure the shared information between predicted partition $X$ and the truth $Y$. 
$NMI$ is defined as $NMI(X,Y) = 2\frac{{I(X;Y)}}{{H(X) + H(Y)}}$, where $I$ is mutual Information and $H$ is entropy. $NMI \in \left[ {0,1} \right]$. 
When $X$ is more similar with $Y$,  $NMI$ is closer to 1.
As our only hyper-parameter, $\alpha$ is set to 1.

\noindent \textbf{$\bullet$ Baselines. }
Our model is compared to baselines of topic representations and clustering methods. (1) As different topic representations, we compared to Latent Dirichlet Allocation (LDA) \cite{blei2003latent}, TF-IDF, and mean-pooled token/word vector of BERT \cite{devlin2018bert} and GloVe \cite{pennington2014glove} embedding (Line 1-4). We also compare with the combinations of embeddings with SIF or SAE (Line 5-10).
(2) As a naive clustering methods, the plain $k$-means is used as a baseline as well (Line 2-10).

\subsubsection{Results}
Results show that our model outperforms all baselines.
Metrics of all settings are shown in Table \ref{cluster results}, which is presented in three parts.
The first and second parts includes baselines on $k$-means for clustering, while the second part adds SIF or SAE.
The third part includes results of our model. 

As shown in the first part of Table  \ref{cluster results} (Line 1-4), LDA has fewer topic clusters, higher $C_{rate}$  but lower $A_{rate}$, which means it tends to produce large but mixed clusters. Those clusters would have low accuracy and become confusing.
TF-IDF in English dataset behaves better than GloVe, which can be attributed to frequent and informative words such as "$hotel$", "$restaurant$". 
Compared with other encoding methods, 
BERT performs better in almost all metrics due to contextualized representations.

In the second part (Line 5-10), we introduce SIF and SAE for further encoding and reducing dimensions. 
Applying SIF, GloVE+SIF (Line 5) and BERT+SIF (Line 6) all achieves better results on both topic numbers $N_c$ and accurate coverage rate $A_{rate}$ which is our most concerned metric, compared to GloVE (Line 3) and BERT (Line 4). These comparisons generally show that SIF is strong enhancement for text representations especially for our concerned topic clustering task. 
Then SAE is applied to GloVE and BERT with or without SIF (Line 7-10). 
Comparing line 8,10 with 6, and line 7,9 with 5, we can tell that SAE leads to compromised performance, which may be caused by inappropriate dimensions reducing.

The third part (Line 11-14) shows that our method of BERT+SIF on SAE+Self-training achieves the best results on almost all metrics of both datasets, especially the most important $A_{rate}$.
Our method can cover most topics while make clusters most accurate among all methods.
The clustering method of SAE+Self-training performs much better than simple $k$-means. Because adjustments are made based on the clusters of $k$-means, and our BERT+SIF encoding method further improves on that. 

\begin{figure}[ht]
\centering
\includegraphics[scale=0.28]{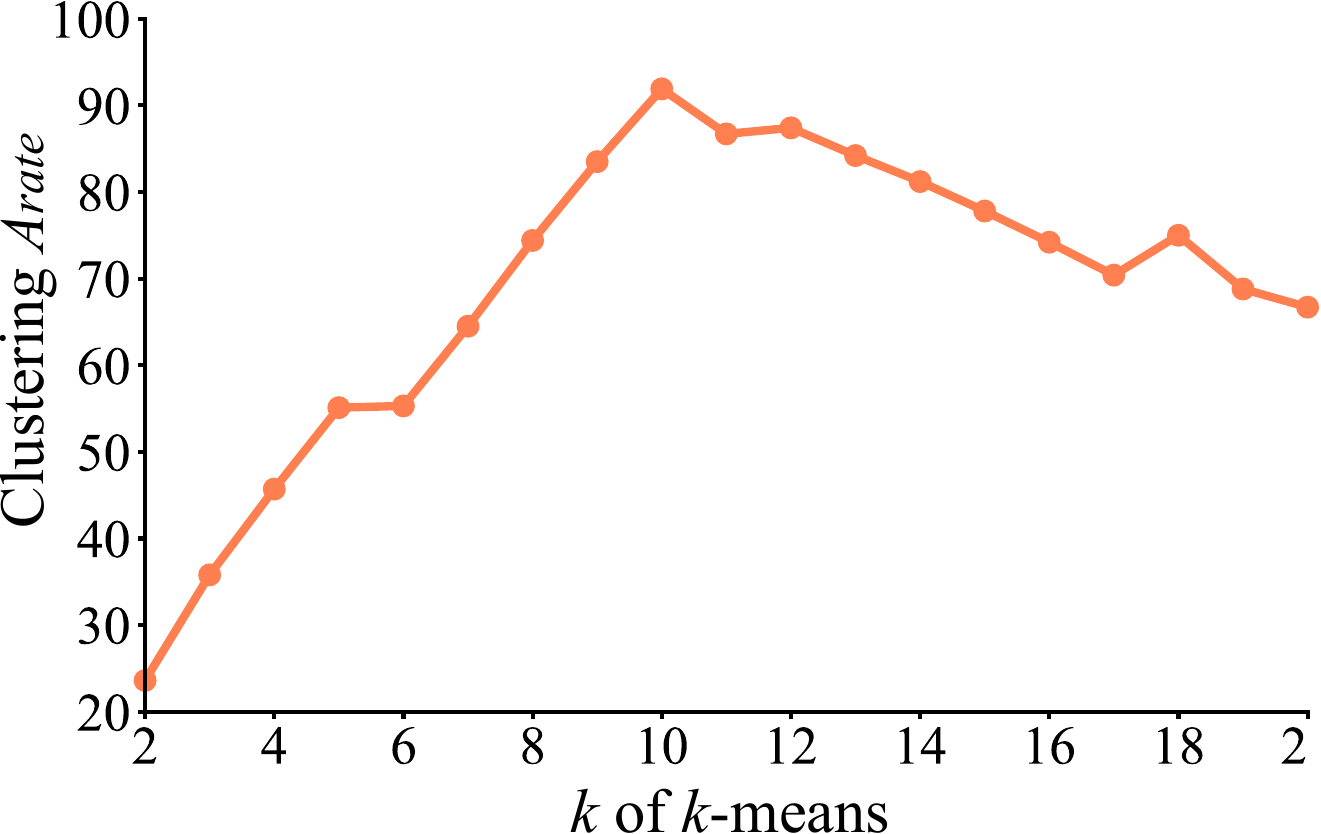}
\caption{Illustration of Clustering $A_{rate}$ under different value $k$ for $k$-means in English dataset.}
\label{ablation}
\end{figure}
\subsubsection{Analysis}
\textbf{$\bullet$ Number of Clusters.}
We also explore the influence of the number of clusters, that is, value $k$ for $k$-means in Figure \ref{ablation} in English dataset. Clustering $A_{rate}$ increases sharply with $k$ until reaching 10, which is true cluster number of the dataset. Then $A_{rate}$ begins to drop. Value of $k$ is of great influence, especially when $k$ is smaller than the true cluster number.

\noindent \textbf{$\bullet$ End-to-end Clustering.}
\begin{table}
\centering
\renewcommand\tabcolsep{8pt}
\caption{\label{diatance measuring tabel} An end-to-end clustering results vs. stepwise results. }
\begin{tabular}{lcccc}
\hline 
\textbf{Dataset}&$N_c$&Segmentation&Clustering&end-to-end \\ 
&&$F_1$ & $A_{rate}$(\%) &$F_1^{all}$\\ 
\hline
Chinese&6&0.61&42.5&0.33\\
English&10&0.61&91.9&0.55\\
\hline
\end{tabular}
\end{table}
\begin{figure}[ht]
\centering
\includegraphics[scale=0.25]{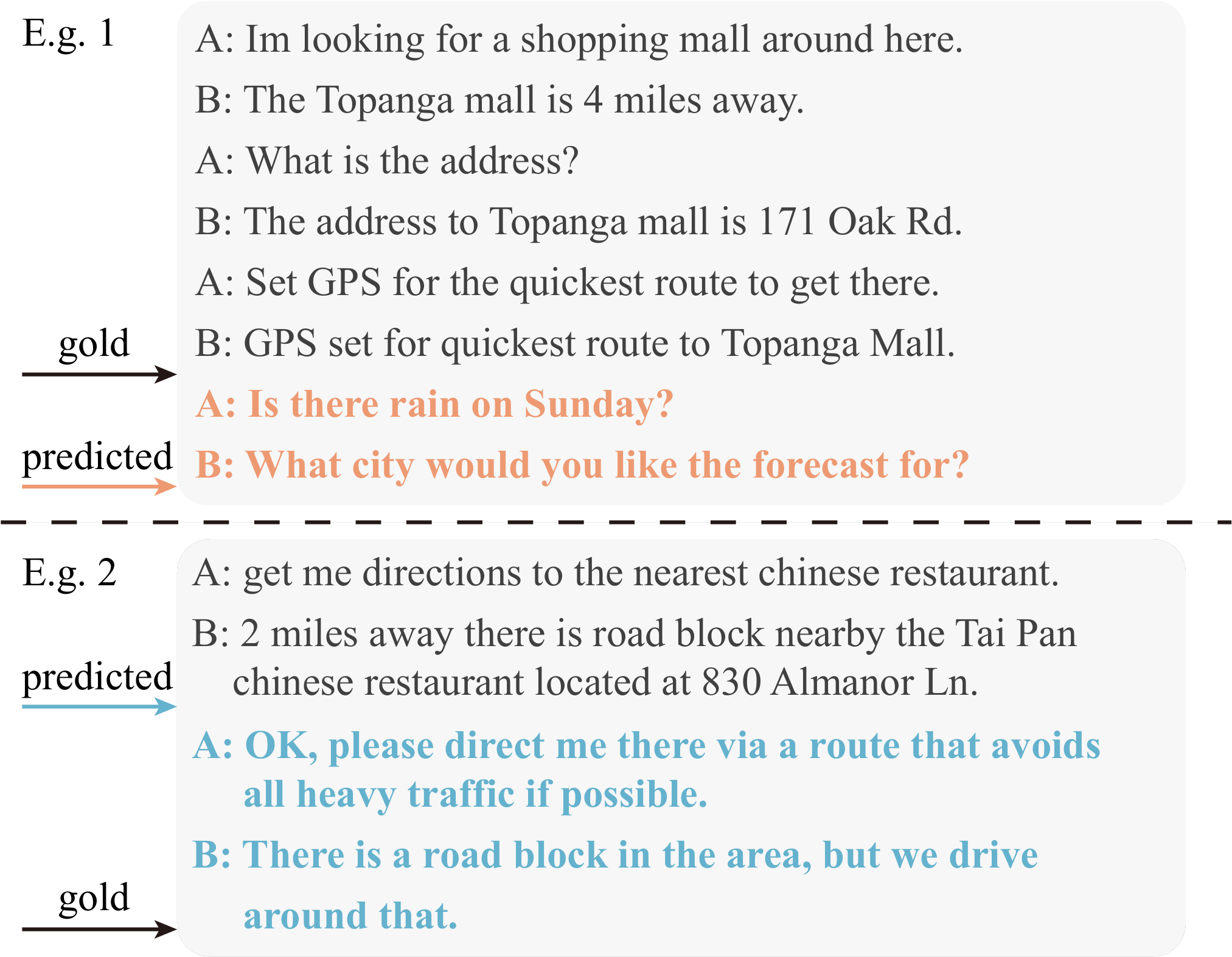}
\caption{\label{cluster example tabel} End-to-end clustering examples. The black text is the overlapping, and colored text are redundant or missing utterances.}
\end{figure}
\label{cluster examples}
\begin{table*}[ht]
\centering
\caption{\label{tab: sel_results} Multi-turn dialogue topic-aware response selection results on Ubuntu, Douban and E-commerce datasets. "\dag" means topic-related work.}
\begin{tabular}{lccc|cccccc|ccc}
\toprule 
\multirow{2}*{\textbf{Model}}& \multicolumn{3}{c}{Ubuntu}&\multicolumn{6}{c}{Douban}& \multicolumn{3}{c}{E-commerce}\\ 
~&$R_{10}@1$&$R_{10}@2$&$R_{10}@5$&MAP&MRR&P@1&$R_{10}@1$&$R_{10}@2$&$R_{10}@5$&$R_{10}@1$&$R_{10}@2$&$R_{10}@5$\\ 
\midrule 
TF-IDF \cite{lowe-etal-2015-ubuntu}& 41.0 &54.5& 70.8&33.1 &35.9& 18.0& 9.6 &17.2& 40.5&15.9&25.6&47.7 \\
RNN \cite{lowe-etal-2015-ubuntu}&40.3& 54.7& 81.9&39.0& 42.2& 20.8& 11.8 &22.3& 58.9&32.5&46.3&77.5\\
CNN \cite{kadlec2015improved}&54.9& 68.4& 89.6&41.7& 44.0 &22.6& 12.1 &25.2& 64.7&32.8&51.5&79.2\\
LSTM \cite{kadlec2015improved}&63.8 &78.4& 94.9&48.5 &53.7& 32.0 &18.7& 34.3& 72.0&36.5&53.6&82.8\\
BiLSTM \cite{kadlec2015improved}&63.0 &78.0& 94.4&47.9& 51.4 &31.3& 18.4 &33.0 &71.6&35.5&52.5&82.5\\
DL2R \cite{10.1145/2911451.2911542}&62.6 &78.3& 94.4&48.8& 52.7& 33.0 &19.3 &34.2& 70.5&39.9&57.1&84.2\\
Atten-LSTM \cite{tan2015lstm}&63.3 &78.9& 94.3&49.5& 52.3 &33.1 &19.2& 32.8& 71.8&40.1&58.1&84.9\\
MV-LSTM \cite{10.5555/3060832.3061030}&65.3& 80.4 &94.6&49.8& 53.8& 34.8 &20.2 &35.1& 71.0&41.2&59.1&85.7\\
Match-LSTM \cite{wang-jiang-2016-learning}&65.3& 79.9& 94.4&50.0 &53.7& 34.5 &20.2 &34.8& 72.0&41.0&59.0&85.8\\
\midrule
Multi-View \cite{zhou2016multi}&66.2& 80.1& 95.1&50.5& 54.3& 34.2& 20.2 &35.0& 72.9&42.1&60.1&86.1\\
SMN \cite{wu-etal-2017-sequential}&72.6 &84.7& 96.1&52.9 &56.9& 39.7& 23.3& 39.6& 72.4&45.3&65.4&88.6\\
DUA \cite{zhang-etal-2018-modeling}&75.2&86.8& 96.2&55.1 &59.9& 42.1& 24.3& 42.1& 78.0&50.1&70.0&92.1\\
DAM \cite{zhou2018multi}&76.7& 87.4& 96.9&55.0& 60.1& 42.7 &25.4& 41.0& 75.7&-&-&-\\
MRFN\cite{tao2019multi}&78.6& 88.6& 97.6&57.1&61.7& 44.8& 27.6& 43.5& 78.3
&-&-&-\\     
IMN \cite{gu2019interactive}&79.4&88.9&97.4&57.0& 61.5& 44.3 &26.2& 45.2& 78.9&62.1&79.7&96.4\\
IOI \cite{tao-etal-2019-one}&79.6&89.4&97.4&57.3& 62.1& 44.4 &26.9 &45.1 &78.6&56.3&76.8&95.0\\
MSN \cite{yuan2019multi}&80.0&89.9&97.8&58.7& 63.2& \textbf{47.0} &\textbf{29.5}& 45.2 &78.8&60.6&77.0&93.7\\
\midrule
TACNTN\dag \cite{wu2018response} &38.3 &54.4 & 80.9&-&-&-&-&-&-&-&-&-\\
KEHNN\dag \cite{wu2018knowledge} &46.0 & 59.1 & 81.9 &-&-&-&-&-&-&-&-&-\\
\midrule
BERT (Our Baseline)&81.9&90.4&97.8&58.7&62.7&45.1&27.6&45.8&82.7&62.7&82.2&96.2\\
TADAM &\textbf{82.1}&\textbf{90.6}&\textbf{97.8}&\textbf{59.4}&\textbf{63.3}&45.3&28.2&\textbf{47.2}&\textbf{82.8}&\textbf{66.0}&\textbf{83.4}&\textbf{97.5}\\
\bottomrule
\end{tabular}

\end{table*}

To evaluate the joint effect of our segmentation and clustering algorithms in an end-to-end situation, we use metrics $F_{1}^{all}$ 
to measure the distance between golden clusters and predicted clusters. 
For a dialogue, we first match each predicted topic segment with the golden one based on $F_1$ calculated by their overlap range of utterances, and we set an overlapping threshold for $F_1$ (here is 50\%) as filter. Then we use Hungarian Algorithm to predict the topics of segments. 
$F_{1}^{all}$ is calculated as the mean $F_1$ of all clusters. 

Table \ref{diatance measuring tabel} lists the end-to-end clustering results. We use BERT$_{CLS}$ and BERT$_{mean}$ in our topic-aware segmentation algorithm for Chinese and English datasets respectively, and use BERT+SIF with SAE+Self-training for topic clustering on both datasets. We can find that our proposed algorithms jointly give good enough, practical results even they suffer more or less from the error-propagation of two-stage processing. 
Figure \ref{cluster example tabel} gives two bad cases. On clustering stage, they both get correct prediction of the topic \textit{Navigate}, but the transition points are biased caused by the segmentation stage. The black texts are overlapping of the predicted and gold segmentation. But the transition point of example 1 is two utterance later, mixed with weather-related sentences. While the transition point of example 2 is earlier, and the missing utterances are about traffic, which is a little different but still related.

\subsection{Experiment \uppercase\expandafter{\romannumeral3}: Response Selection}
\label{rslab}
This section shows experiments of TADAM model on response selection.
Inheriting the segmentation algorithm, resulting topic segments are used as input units of a dialogue passage.
Ablation studies are shown to verify contribution of TADAM parts. And we show solid evidence on the effect of processing units.
\subsubsection{Dataset}
\begin{table}[tbh]
  \renewcommand\tabcolsep{1pt}
\centering
\caption{\label{mydataset_resel} Dataset statistics for response selection.}
\begin{tabular}{l|ccc|ccc|ccc}
\hline 
\multirow{2}*{\textbf{Dataset Statistics}}& \multicolumn{3}{c|}{Ubuntu}&\multicolumn{3}{c|}{Douban}&\multicolumn{3}{c}{E-commerce} \\ 
~&Train&Valid&Test&Train&Valid&Test&Train&Valid&Test\\ 
\hline 
\#Context-response &1M& 500K&500K&1M& 50K& 50K &1M&10K&10K \\
\#Candidates/Context &2&10&10& 2&2&10& 2&2&10 \\
Avg. Turns/Context &10.13&10.11&10.11&6.69&6.75&6.45&5.51&5.48&5.64\\
Avg. Tokens/Utterance&11.35&11.34&11.37&18.56&18.50&20.74&7.02&6.99&7.11\\
\hline
\end{tabular}
\end{table}
For topic-aware response selection, TADAM is tested on three widely used benchmarks. (1) \textbf{Ubuntu Corpus} \cite{lowe-etal-2015-ubuntu}: It consists of English multi-turn conversations about technical support collected from chat logs of the Ubuntu forum. It contains 1 million context-response pairs for training and 0.5 million pairs for validation and testing. The positive-to-negative ratio is 1:1 in training 1:9 in validation and testing. 
(2) \textbf{Douban Corpus} \cite{wu-etal-2017-sequential}: It consists of multi-turn conversations from the Douban group
, a popular social networking service in China. It contains 1 million context-response pairs for training, 50,000 million pairs for validation and testing. The positive-to-negative ratio is 1:1 in training and validation and 1:9 in testing.
(3) \textbf{E-commerce Corpus} \cite{zhang-etal-2018-modeling}: It includes conversations between customers and sellers from the largest e-commerce platform Taobao 
in China. The E-commerce Corpus has an obvious topic shift, including commodity consultation, express, recommendation, and chitchat.  It contains 1 million context-response pairs for training and 10,000 pairs for both validation and testing. The positive-to-negative ratio is 1:1 in training and validation and 1:9 in testing. More details are shown in Table \ref{mydataset_resel}.

\subsubsection{Settings}
\textbf{$\bullet$ Metrics.}
For response selection, we use the same metric $R_n@k$ as previous works, which selects $k$ best matchable candidate responses among $n$ and calculates the recall of the true ones. Besides, because Douban corpus has more than one correct candidate response, we also use MAP (Mean Average Precision), MRR (Mean Reciprocal Rank), and Precision-at-one P@1 as previous works.

\noindent \textbf{$\bullet$ Hyper-Parameters.}
The encoder we use is pre-trained BERT (bert-base-uncased \& bert-base-chinese) here. 
In segmentation stage, we set $R=2, 2, 6$ for Ubuntu, Douban and E-commerce after trying different values. 
As to TADAM, the maximum input sequence length is set to 350 after WordPiece tokenization and the maximum number of segments is 10. We set the learning rate as 2e-5 using BertAdam with a warmup proportion of 10\%. Our model is trained
with batch size of \{20,32,20\} and epoch of \{3,3,4\} for the three benchmarks.  Besides, the $\beta$ in equation \ref{beta} is set to 0.5. 

\noindent \textbf{$\bullet$ Baslines. }
We concatenate the context and candidate response as input for BERT as a basic sequence classification baseline.
The fine tuned epochs are \{3,2,3\} for three datasets while other settings keeps identical with TADAM. 
We also compared with numbers of public baselines including previous works (e.g. LSTM-based models \cite{kadlec2015improved}) and more recent ones (e.g. MSN \cite{yuan2019multi}, IOI \cite{tao-etal-2019-one}).
\subsubsection{Results}
Experimental results in Table \ref{tab: sel_results} show that our model outperforms all public works and especially gets much improvement (3.3\% in $R_{10}@1$) over the strong pre-trained contextualized language model in E-commerce dataset, which shows the effectiveness of our topic-aware models in dialogues with topic shifting scenes. Through observation, we find that the fact of topic shift is negligible in Ubuntu and Douban where a whole dialogue is almost about one topic. This is why improvement of the model in Douban is not as obvious as that in E-commerce with multiple topics. However, our work is not supposed to work best in all scenarios, but especially focuses on the case of topic shift which are common in the more challenging situations like e-commerce or banking. Results show that it does work in specific application scenarios, which right verifies the motivation of this work.

\begin{table}[ht]
\renewcommand\tabcolsep{1.6pt}
\centering
\caption{\label{tab: ablation} Ablation study of TADAM on E-commerce and Ubuntu.}
\begin{tabular}{lcccccc}
\toprule  \small
\multirow{2}*{\textbf{Model}}& \multicolumn{3}{c}{E-commerce}&\multicolumn{3}{c}{Ubuntu}\\ 
~&$R_{10}@1$&$R_{10}@2$&$R_{10}@5$&$R_{10}@1$&$R_{10}@2$&$R_{10}@5$\\ 
\midrule
TADAM&66.0&83.4&97.5&82.1&90.6&97.8\\
\midrule 
w/o word weights&62.0&82.2&96.3&81.9&90.6&97.8\\
w/o seg. weights&63.1&82.7&97.0&81.8&90.5&97.8\\
w/o weights&62.2&82.2&97.5&81.9&90.6&97.7\\
\midrule
w/o last seg. match&64.0&82.9&96.7&82.0&90.5&97.8\\
w/o multi-turn match&63.8&82.5&96.4&81.9&90.5&97.8\\
\midrule
single attention (seg.)&62.7&83.3&97.4&81.9&90.6&97.8\\
single attention (res.)&62.4&83.2&97.6&81.6&90.3&97.8\\
\bottomrule
\end{tabular}
\end{table}
\subsubsection{Analysis}
\textbf{$\bullet$ Ablation Studies.}
In order to investigate the performance of each part of our model, we conduct serious ablation experiments from three angles and results are shown in Table \ref{tab: ablation}. First, we explore the influence of the segment weighting part by removing word or segment level weights or both of them (Line 3-5). Second, in the aggregation part, we concatenate the multi-turn matching result $\hat{H}$ and last segment matching result $\hat C_T^{3}$ to get a score. Hence we remove either of both each time (Line 6-7). 
Third, we do dual cross-attention matching between context segments and response candidates, i.e. Equation \ref{euq:C_S_1} \ref{euq:C_R_1}. So we ablate each attention in Line 8-9.

As shown in Table \ref{tab: ablation}, for E-commerce, removing word or segment level weights all perform worse. Besides, enhancing extra last segment match does make sense as the traditional multi-turn matching method with GRU. 
Moreover, both single attentions lead to much decrease, showing the significance of the dual matching design. 
Results of Ubuntu are not so obvious as that of E-commerce, which can be attributed to that our work especially focuses on the case of topic shift but dialogues in Ubuntu are almost about consistent topic.

\begin{figure}[ht]
\centering
\includegraphics[scale=0.33]{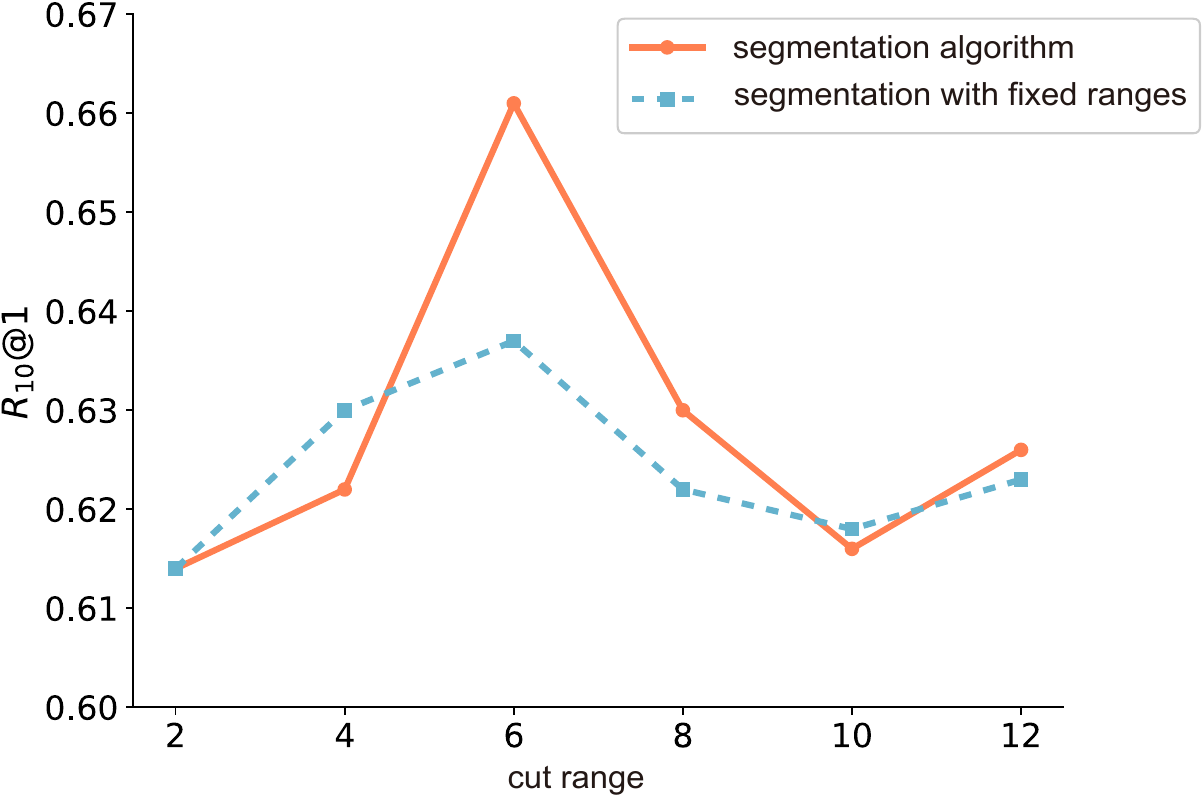}
\caption{$R_{10}@1$ for different ranges on E-commerce.}
\label{fig: fixed_interval_E}
\end{figure}

\noindent \textbf{$\bullet$ Effect of Segmentation. }
In order to investigate the effectiveness of our segmentation algorithm in the response selection task, we use a naive method that simply segment by a fixed length $\hat{R}$. 
$\hat{R}$ is adjusted in the same pace with hyper-parameter $R$ of the segmentation algorithm as comparison.
Results are shown in Figure \ref{fig: fixed_interval_E}. For both methods, with the increase of cut range, range of 6 performs best. Both too small and large intervals hurt performance. Besides, applying our segmentation algorithm performs better than just using fixed ranges in most ranges especially in range of 6, proving the effectiveness of our proposed topic-aware segmentation algorithm.

In addition, this experiment also shows that segmentation performance has a non-negligible impact to the response selection recall, which means error propagation may suppress TADAM performance to some extent.
As there is still a latent room to improve the segmentation results, TADAM has potential on a superior segmentation method.

\begin{figure}[ht]
\centering
\includegraphics[scale=0.3]{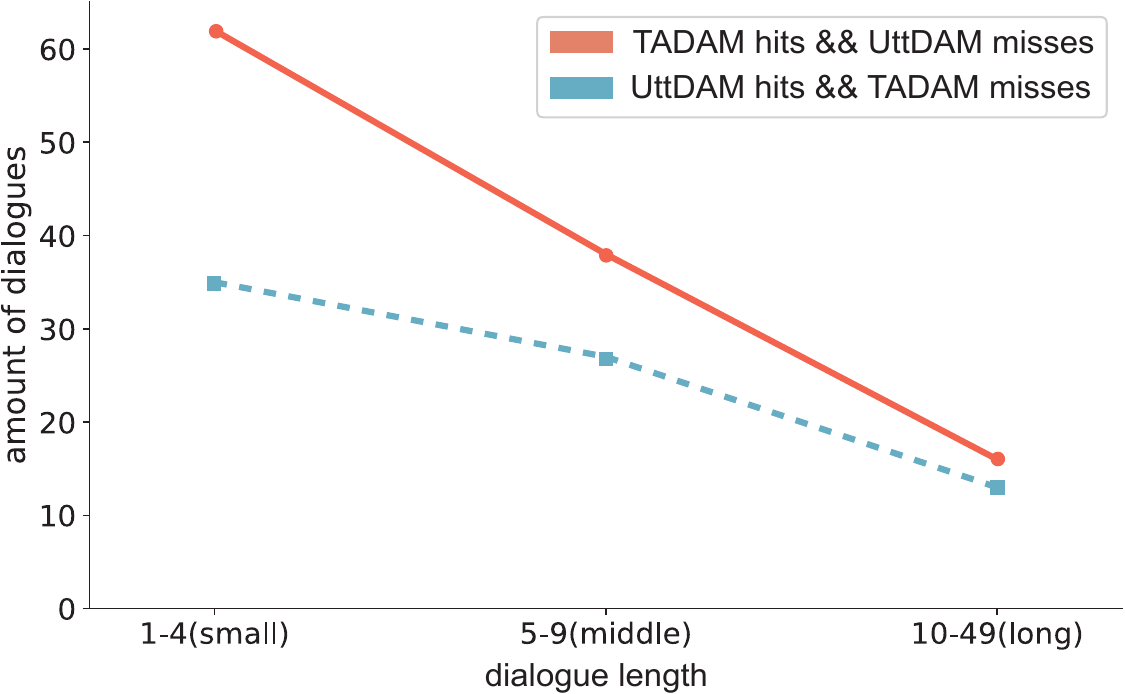}
\caption{Performance of UttDAM and TADAm on dialogue length distribution.}
\label{fig: dialogue len distribution}
\end{figure}

\noindent \textbf{$\bullet$ Effects of Input Units.}
\label{sec: topic_seg_effect}
\begin{table}[ht]
 \centering
 \caption{\label{tab: input} Results of different input units.}
 \renewcommand\tabcolsep{12pt}
\begin{tabular}{lccc}
\toprule 
\textbf{Model}&$R_{10}@1$&$R_{10}@2$&$R_{10}@5$\\ 
\midrule
TADAM&66.0&83.4&97.5\\
\midrule
UttDAM&62.9&81.5&97.2\\
separated segment &36.0&49.2&72.5\\
\bottomrule
\end{tabular}
\end{table}
To verify the necessity of topic-aware segmentation in the response selection task, we remove the segmentation part and just concatenate utterances as well as response.
In this case, the subsequent modules are all based on utterance units, which is noted as UttDAM.
Its results in line 3 of Table \ref{tab: input} decrease by a large margin, which implies that taking the segment as a unit is more robust to the irrelevant contextual information than the utterance.
Figure \ref{fig: dialogue len distribution} shows performances of TADAM and UttDAM on dialogues with different length. We can observe that segment units outperform on all length but more on shorter ones.

\begin{table}[ht]
    \centering
        \caption{ \label{tab:dialogue_case_2}A topic-aware segmentation case from E-commerce Corpus.}
    \begin{tabular}{|c|l|}
    \hline
           \textbf{Turns} & \multicolumn{1}{c|}{\textbf{Dialogue Text}} \\
           \hline
           Turn-1 & A: \textit{Hello.} \\
        Turn-2 & B: \textit{Excuse me, has my order been sent out?} \\
        \arrayrulecolor{black}\hdashline
        Turn-3 & A: \textit{Please let me check.} \\
        \arrayrulecolor{black}\hline
        Turn-4 & B: \textit{I found I didn't buy the cotton one.} \\
        Turn-5 & A: \textit{Your order has been sent out.} \\
        Turn-6 & B: \textit{It's non-woven fabric.}\\
        Turn-7 & A: \textit{Yes.} \\
        Turn-8 & B: \textit{I'd like to switch to the plant fiber.} \\
        \arrayrulecolor{black}\hdashline
        Turn-9 & A: \textit{Ok.} \\
        \arrayrulecolor{black}\hline
        Turn-10 & B: \textit{Please change it for me.} \\
        Turn-11 & A: \textit{Sorry, your order has been taken by the courier.} \\
        Turn-12 & B: \textit{Can you get it back?} \\
        Turn-13 & A: \textit{I'll try to intercept for you} \\
        Turn-14 & B: \textit{I'm sorry} \\
        \arrayrulecolor{black}\hdashline
        Turn-15 & A: \textit{It doesn't matter} \\
        \arrayrulecolor{black}\hline
        Turn-16 & B: \textit{What is the natural plant fiber?} \\
        \arrayrulecolor{black}\Xhline{0.8pt}
    \end{tabular}
    \smallskip
	\\ \footnotesize{Note: Solid lines are right boundaries, and dotted lines are labelled by our segmentation algorithm.}
\end{table}

we also explore the encoding mode for topic segments. In TADAM, topic segments and candidate response are concatenated to feed into the encoder, and then are split for further matching. 
Here we encode each segment and response separately, indicating that the segment or response itself just focuses on its own meaning without interaction with other contexts. Results in line 4 in Table \ref{tab: input} show that this encoding mode causes a heavy performance loss. It means that although the segment itself can be encoded purely, it leads to information scarcity and can be more sensitive to segmentation error. 
We explain this phenomenon with Table \ref{tab:dialogue_case_2} that shows a topic-aware segmentation case from E-commerce Corpus. We can find that our algorithm does split out topic segments, whose boundaries are near to the true ones. Nevertheless, it is not ideal to encode segments and response separately. Encoding concatenated segments allows for relevant information supplement, while separated encoding tends to suffer from mixed topics caused by segmentation deviation.

\section{Conclusion}
This paper proposes to model multi-turn dialogues from a topic-aware perspective, in terms of explicitly segmenting and extracting topic-aware segments for the dialogue comprehension tasks. 
An effective topic-aware segmentation algorithm is well designed to generate topic segments as processing units for two practices in multi-turn dialogues, hot topic detection and response selection. 
For hot topic detection,
we pre-train an autoencoder for segment encoding and further use self-training to improve the clustering.
To fulfill our research purpose of topic detection, we build two novel datasets with annotations of topic transition points and topic labels, which is the first topic transition dataset. 
For the response selection, we propose a topic-aware dual-attention network TADAM, to further improve the matching of response and segmented contexts. Experimental results of TADAM on three benchmarks show the significance. In conclusion, this work propose to model dialogue with awareness of topic and presents solid empirical studies, providing a new perspective for dialogue reading comprehension.


%

%



\ifCLASSOPTIONcaptionsoff
  \newpage
\fi



\bibliographystyle{IEEEtran}
\bibliography{reference}
\end{document}